\documentclass[10pt,twocolumn,letterpaper]{article}

\usepackage{wacv}
\usepackage{times}
\usepackage{epsfig}
\usepackage{graphicx}
\usepackage{amsmath}
\usepackage{amssymb}
\usepackage{booktabs}
\usepackage{subcaption}
\usepackage{color}

\usepackage{multirow}
\usepackage{xspace}
\usepackage{booktabs,siunitx}
\usepackage{colortbl}
\definecolor{mygray}{gray}{.9}

\newcommand{\tablestyle}[2]{\setlength{\tabcolsep}{#1}\renewcommand{\arraystretch}{#2}\centering\footnotesize}

\usepackage{color}
\usepackage[pagebackref=true,breaklinks=true,colorlinks,urlcolor={gray},citecolor=Green, bookmarks=false]{hyperref}
\usepackage[capitalize]{cleveref}
\crefname{section}{Sec.}{Secs.}
\Crefname{section}{Section}{Sections}
\Crefname{table}{Table}{Tables}
\crefname{table}{Tab.}{Tabs.}
\usepackage{bbding}
\usepackage{bibunits}
\usepackage{bm}
\usepackage{diagbox}
\usepackage{array}
\usepackage[usenames,dvipsnames]{xcolor}
\usepackage{tabularx}
\usepackage{booktabs}
\usepackage{pdflscape}
\usepackage{makecell}
\usepackage{framed,multirow}
\usepackage[flushleft]{threeparttable}
\usepackage{amssymb}% http://ctan.org/pkg/amssymb
\usepackage{pifont}% http://ctan.org/pkg/pifont
\usepackage{stfloats}
\usepackage{algorithm}
\usepackage{layouts}
\usepackage{listings}
\usepackage{pythonhighlight}
\hypersetup{
    colorlinks=true,
    citecolor=green,
    urlcolor=blue,
}
\makeatletter
\newcommand\footnoteref[1]{\protected@xdef\@thefnmark{\ref{#1}}\@footnotemark}
\makeatother

\newcolumntype{P}[1]{>{\centering\arraybackslash}p{#1}}
\newlength\savewidth\newcommand\shline{\noalign{\global\savewidth\arrayrulewidth
  \global\arrayrulewidth 0.8pt}\hline\noalign{\global\arrayrulewidth\savewidth}}
\newcommand{\cmark}{\ding{51}}%
\newcommand{\xmark}{\ding{55}}%
\definecolor{deemph}{gray}{0.6}
\newcommand{\gc}[1]{\textcolor{deemph}{#1}}

\newcolumntype{x}[1]{>{\centering\arraybackslash}p{#1pt}}
\newcolumntype{y}[1]{>{\raggedright\arraybackslash}p{#1pt}}
\newcolumntype{z}[1]{>{\raggedleft\arraybackslash}p{#1pt}}

\definecolor{deemph}{gray}{0.6}
\definecolor{baselinecolor}{gray}{.9}

\newenvironment{jlmagenta}{\color{black}}{}

\wacvfinalcopy % *** Uncomment this line for the final submission

\ifwacvfinal
\usepackage[breaklinks=true,bookmarks=false]{hyperref}
\else
\usepackage[pagebackref=true,breaklinks=true,colorlinks,bookmarks=false]{hyperref}
\fi

\begin{document}

%%%%%%%%% TITLE

\title{Delving into Masked Autoencoders for Multi-Label Thorax Disease Classification}

\author{Junfei~Xiao~~~~~~~~Yutong~Bai~~~~~~~~Alan~Yuille~~~~~~~~Zongwei~Zhou\thanks{Corresponding author: Zongwei Zhou (\href{mailto:zzhou82@jh.edu}{zzhou82@jh.edu})}\\[2mm]
Johns Hopkins University\\[1mm]
{\small Code: \href{https://github.com/lambert-x/Medical_MAE}{https://github.com/lambert-x/Medical\_MAE}}
}

\maketitle

\begin{abstract}

Vision Transformer (ViT) has become one of the most popular neural architectures due to its great scalability, computational efficiency, and compelling performance in many vision tasks. However, ViT has shown inferior performance to Convolutional Neural Network (CNN) on medical tasks due to its data-hungry nature and the lack of annotated medical data.
In this paper, we pre-train ViTs on 266,340 chest X-rays using Masked Autoencoders (MAE) which reconstruct missing pixels from a small part of each image.
For comparison, CNNs are also pre-trained on the same 266,340 X-rays using advanced self-supervised methods (\eg MoCo~v2). 
The results show that our pre-trained ViT performs comparably (sometimes better) to the state-of-the-art CNN (DenseNet-121) for multi-label thorax disease classification. 
This performance is attributed to the strong recipes extracted from our empirical studies for pre-training and fine-tuning ViT.
The pre-training recipe signifies that medical reconstruction requires a much smaller proportion of an image (10\%~vs.~25\%) and a more moderate random resized crop range (0.5$\sim$1.0~vs.~0.2$\sim$1.0) compared with natural imaging. Furthermore, we remark that in-domain transfer learning is preferred whenever possible.
The fine-tuning recipe discloses that layer-wise LR decay, RandAug magnitude, and DropPath rate are significant factors to consider.
We hope that this study can direct future research on the application of Transformers to a larger variety of medical imaging tasks.

\end{abstract}
\section{Introduction}
\label{sec:introduction}

There has been great progress in the Vision Transformer (ViT) architecture~\cite{dosovitskiy2020image} and its variants~\cite{liu2021swin,hassani2021escaping,guo2022beyond,touvron2021training}, showing that Transformers surpass and supersede Convolutional Neural Networks (CNN) in various natural imaging tasks. In comparison with CNNs, Transformers can better leverage the rapidly increasing image data, long-range spatial context of an image~\cite{cordonnier2019relationship,ding2022scaling}, and share properties of the human visual system~\cite{naseer2021intriguing,portelance2021emergence,geirhos2021partial,tuli2021convolutional}.
Training Transformers requires considerably more data than CNNs~\cite{steiner2021train,tay2022scaling}, but medical data are small and labels are hard to obtain.
As a result, directly applying Transformers to the medical domain is found to be problematic and challenging.
There are several early attempts~\cite{matsoukas2021time,xie2021unified,ma2022benchmarking}, but their performance is often inferior to state-of-the-art CNNs (elaborated in \S\ref{sec:related_work}). Recent surveys suggest that a range of successful cases are using a hybrid architecture of Transformers and CNNs~\cite{li2022transforming,shamshad2022transformers}. 
In contrast, the \textit{stand-alone} and \textit{vanilla} ViT architecture remains the concentration of this study to strive for simplicity.
We ask: \textit{What is the full potential of ViT architecture in medical imaging tasks?} 
The answer, based on our study, is that vanilla ViT can achieve a similar or even better performance than state-of-the-art CNNs if equipped with (I) a large-scale pre-training on unlabeled medical data and (II) strong pre-training and fine-tuning recipes, customized by unique characteristics of medical images.

The pre-training of CNNs has been widely investigated in the medical domain~\cite{hosseinzadeh2021systematic}, resulting in several publicly available Foundation models~\cite{zhou2021models,chen2019med3d,xie2021unified}. Numerous pre-training methods can enable CNNs to learn representation from unlabeled images, including contrastive learning~\cite{taher2022caid}, predictive learning~\cite{zhu2020rubik}, restorative learning~\cite{chen2019self}, and their combination~\cite{guo2022discriminative,haghighi2022dira}. At the time this paper is written, however, neither contrastive nor predictive pre-training is mature for vanilla ViT architectures yet. The most popular pre-training scheme for ViTs is called Masked Autoencoders (MAE)~\cite{he2022masked}. Its task is to mask random patches of the input image and reconstruct the missing pixels. 
We adopt MAE in this paper because of its great scalability, computational efficiency, and compelling performance in many vision tasks.

This paper customizes the recipe of pre-training and fine-tuning MAE for the medical domain and verifies its effectiveness on three chest X-ray datasets. We have also made the pre-training and fine-tuning code publicly available and released ViT-Small and ViT-Base that are pre-trained on 510K X-ray images as well as the pre-trained CNNs. The pre-trained ViT encoder can be fine-tuned to improve classification tasks (validated in \S\ref{sec:finetuning_recipe_results}) and detection tasks (see Github). 
In summary, four contributions are made.

\begin{enumerate}
    \item The usefulness of ViT pre-trained on ImageNet (14M data \& labels) and chest X-rays (0.3M data) is evaluated, underlining the opportunity of in-domain transfer learning and self-supervised learning (\tableautorefname~\ref{tab:comparison_init}).
    \item A strong pre-training recipe, consisting of more unlabeled data (266,340), a higher masking ratio (90\%), and a modest random cropping scale (0.5$\sim$1.0), is developed for MAE to learn image representation from chest X-rays efficiently (\S\ref{sec:more_data}--\ref{sec:less_cropping}).
    \item Three of the most important hyper-parameters are determined to fine-tune ViT in multi-label thorax disease classification: layer-wise LR decay, RandAug magnitude, and DropPath rate (\tableautorefname~\ref{tab:ablations_ft_recipe}).
    \item This is among the first efforts to approach vanilla ViT's performance to the state-of-the-art CNNs on three predominant chest X-ray benchmarks, yielding mAUC of 82.3\%, 89.2\%, 99.3\% on NIH ChestX-ray14, Stanford CheXpert, and COVIDx, respectively (\S\ref{sec:chexpert_result}--\ref{sec:chestxray14_result}).
\end{enumerate}

By intention or non-intention, the empirical comparisons between old and new techniques (\eg CNN~vs.~ViT) are often biased to the newer one~\cite{liu2022convnet,bai2021transformers}. In this paper, we try our best not to over-sell or under-analyze the ViTs' potential in the medical domain. To provide a fair and comprehensive benchmark, the performance of CNNs is truly state-of-the-art in each dataset based on our extensive literature review.
\section{Related Works}
\label{sec:related_work}

\noindent\textbf{\textit{Preliminary.}} 
Radiography images possess unique characteristics compared with photographic images, resulting in considerable difficulties when switching computer vision advancements to medical imaging~\cite{zhou2021towards,zhou2022interpreting,li2022transforming,shamshad2022transformers}. Photographic images, particularly those on ImageNet~\cite{deng2009imagenet}, contain large, apparent objects in the center of the images, residing in varying backgrounds. Learning discriminative features (\eg color, texture, and shape) primarily from the foreground objects is important in computer vision. In contrast, radiography images are generated from pre-defined imaging protocols, so the background exhibits anatomical consistency across images (see examples of the chest anatomy in~\figureautorefname~\ref{fig:grad_cam_small}). 
Clinically relevant information is dispersed throughout the image, whereas the diseased region (as the foreground) often encloses much more local, subtle, and fine-grained variations than photographic images.
As a result, the model must be able to extract both global and local features to identify various diseases from the normal anatomy.
In the following sections, we describe the difference between computer vision and medical imaging in the choice of \textit{model architectures} and \textit{self-supervised methods}, followed by a review of current state-of-the-art solutions for multi-label thorax disease classification.

\smallskip\noindent\textbf{\textit{ViTs or CNNs for medical imaging?}} 
Transformers have gained prevalence in numerous AI applications (\eg AlphaFold2~\cite{jumper2021highly}, Google Translate~\cite{brown2020language}).
In computer vision, there is a heated debate between the adoption of ViTs and CNNs, in terms of performance~\cite{liu2022convnet,zhou2021ibot,bao2021beit,xiao2022transforming,touvron2021training,ding2022scaling}, robustness~\cite{bai2021transformers,mao2022towards,zhang2022delving,zhou2022understanding}, data requirement~\cite{dosovitskiy2020image,steiner2021train,tay2022scaling}, computational efficiency~\cite{paul2022vision}.
This discussion has finally been converging to an agreement that ViTs could serve as alternatives to CNNs in a variety of tasks~\cite{khan2021transformers,han2022survey}.
ViT has substantial potential for radiography imaging tasks, but currently, the superior performance of ViT has not been translated to radiography imaging, where CNN is still the dominant architecture. 
(1) ViTs' performance has lagged behind that of SOTA CNNs~\cite{park2021vision,taslimi2022swinchex}, in which we believe the poorly configured training recipe\footnote{Isensee~\etal~\cite{isensee2021nnu} remark that most of the performance improvement comes for medical imaging is choosing the perfect data process, model training, and optimization strategy of the network (U-Net in their case).} is one of the major causes;
(2) most existing studies report ViTs' performance on medical tasks without comparing with CNN under an \textit{similar} experimental setting~\cite{matsoukas2021time,li2022transforming}; and
(3) multiple works focus on designing \textit{hybrid} architectures by integrating benefits of ViTs and CNNs to claim the superior performance to CNNs~\cite{chen2021transunet,zhang2021transfuse,xie2021cotr,tang2022self}.
Conducting a fair comparison between ViTs and CNNs should take into account the number of parameters, volumes of computations, usages of GPUs, and suitable pre-training schemes. 
So far, there is no broad benchmark to fairly compare ViTs and CNNs in medical tasks, leaving us wondering whether we could trivially switch to ViTs in medical tasks. 
Unlike the aforementioned studies, our objective is to faithfully benchmark between ViTs and SOTA CNNs in radiography imaging tasks; to improve the recipe of existing ViTs with respect to data, model, optimization aspects; and to visualize how ViTs and CNNs interpret radiography images (\S\ref{sec:discussion}).

\smallskip\noindent\textbf{\textit{Self-supervised methods in medical imaging.}} 
Self-supervised learning has shown enormous potential in medical imaging due to the sparsity of high-quality annotation~\cite{zhou2021towards}.
Two major trends are based on contrastive and restorative pre-training. 
In computer vision, contrastive pre-training~\cite{chen2020improved,chen2020simple,grill2020bootstrap,caron2020unsupervised} holds state-of-the-art performance, surpassing supervised ImageNet pre-training in some tasks; while in medical imaging, restorative pre-training~\cite{zhou2021models,tang2022self,feng2020parts2whole} presently reaches a new height in performance.
We attribute this popularity asymmetry to the marked difference between photographic and radiography images. 
Since radiography imaging protocols assess patients in a fairly consistent orientation, the generated images have great similarity across various patients~\cite{xiang2021painting,haghighi2022dira}. 
The inherent consistency eases the analysis of numerous critical problems but also causes a significant problem for contrastive pre-training.
Contrastive pre-training (\eg MoCo~\cite{he2020momentum,chen2020improved}) treats each image as a distinct class and minimizes the similarity of representations derived from different images. This concept, in theory, might not work properly for radiography imaging because the negative pairs appear too similar (empirically evidenced in our~\tableautorefname~\ref{tab:comparison_init}).
In contrast, restorative pre-training is good at conserving fine-grained textures embedded in image context, so it has been widely adopted in medical pre-training.
Restorative pre-training is formulated as the task of pixel-wise image reconstruction~\cite{alex2017semisupervised,chen2019self,zhou2019models,zhu2020rubik,chen2020imagegpt,xie2021simmim}.
Following this spirit, we take masked autoencoders (MAE)~\cite{he2022masked} as pre-training task for its simplicity, efficiency, scalability, and compelling performance.
We are among the first to configure a strong recipe for both ViT pre-training and fine-tuning on enormous chest X-rays. Besides, we extend MAE to pre-train CNNs on the same scale medical data, establish the first direct benchmark between ViTs and SOTA CNNs on public radiography imaging datasets, and extract reusable insights to the medical vision community.

\section{Method}
\label{sec:method}

\smallskip\noindent\textbf{\textit{Data.}} Data from three public X-ray datasets are used to pre-train ViTs (and CNNs as comparison): NIH ChestX-ray14 (75,312 X-rays), Stanford CheXpert (191,028 X-rays), and MIMIC-CXR (243,334 X-rays). All data are in the posteroanterior (PA) or anteroposterior (AP) view, and resized to 256$\times$256 as input. All the X-rays are standardized by mean and standard deviation computed from ImageNet. We perform random resized cropping with a scale range of (0.5$\sim$1.0) and random horizontal flipping. No other data augmentation is applied unless noted.
The pre-training does not require any annotations shipped with the datasets.

\smallskip\noindent\textbf{\textit{Task.}}
The ViT pre-training\footnote{The MAE-style pre-training for CNNs (for comparison in~\tableautorefname~\ref{tab:comparison_init}) is similar to the image in-painting task proposed in Models Genesis~\cite{zhou2021models}.} is analogous to the image reconstruction task proposed in MAE~\cite{he2022masked}: to reconstruct the masked image patches from visible ones.
Mean squared error is computed between the reconstructed and original images in the pixel space, averaged over masked patches~\cite{devlin2018bert}.
 
\begin{figure}[!t]
\centering
    \includegraphics[width=1.0\linewidth]{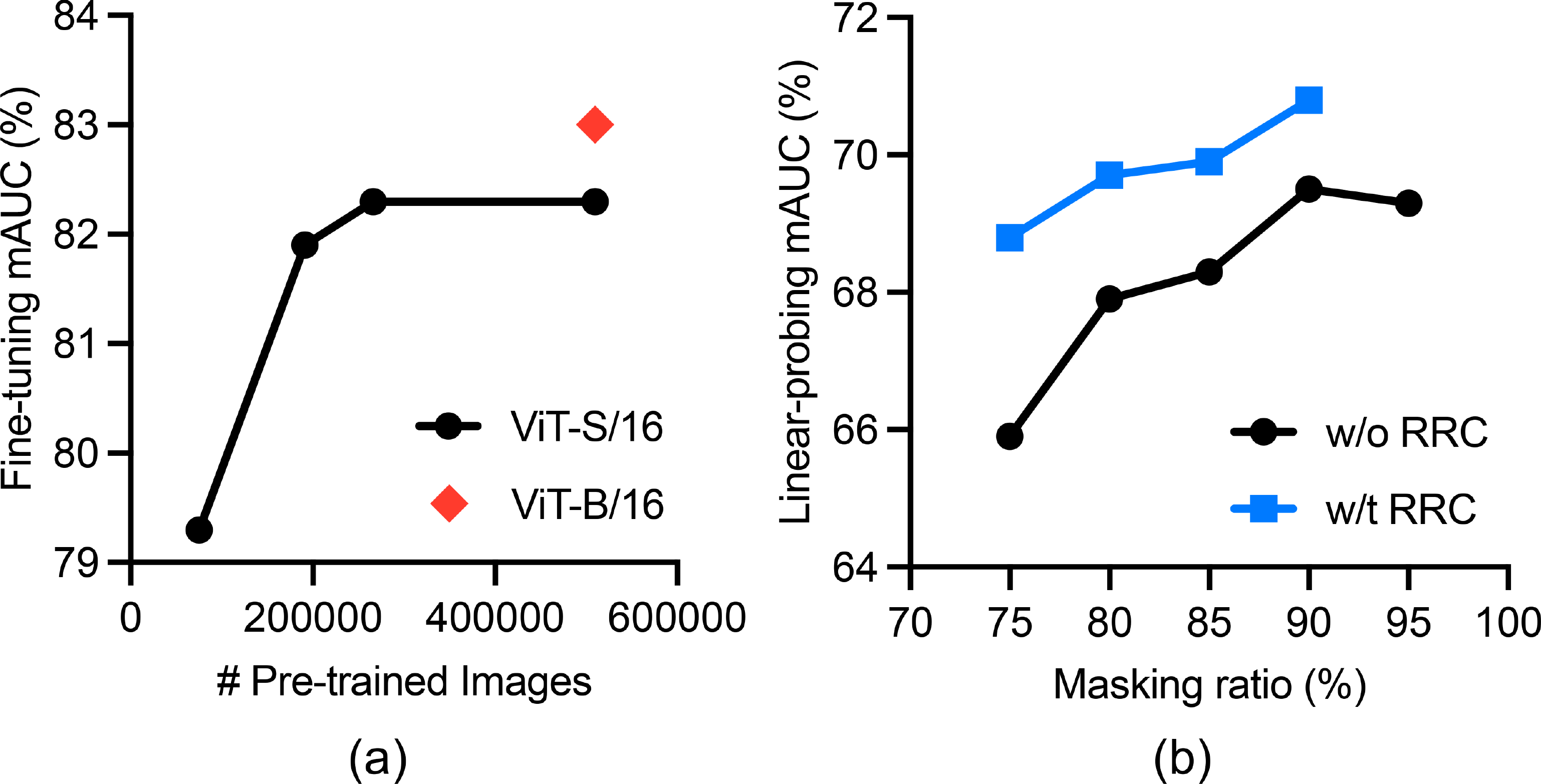}
    \caption{
    \textbf{The pre-training recipe.} (a) Using more images for pre-training can enhance the transferability of ViTs to some extent (\S\ref{sec:more_data}). While ViT-S/16 (Params=22M) seems to be saturated at 266K images, ViT-B/16 (Param=86M) has the potential to scale up to more data. (b) MAE shows the optimal performance at a 90\% masking ratio (\S\ref{sec:more_mask}). Besides, random resized crop (RRC) brings consistent performance gain to MAE pre-training (\S\ref{sec:less_cropping}).
    }
    \label{fig:pretrain_recipe}
\end{figure}

An image is divided into regular non-overlapping patches as a sequence of embeddings.
We randomly sample patches to be masked. The optimal masking ratio we observe is 90\%, which substantially accelerates the pre-training by 2.5$\times$ compared with the original MAE~\cite{he2022masked}\footnote{It should have taken $\sim$16.7 GPU days for the original MAE to pre-train ViT-S/16 (the smallest ViT) on 510K X-rays.} and enables us to scale up ViTs with greater model capability (\figureautorefname~\ref{fig:pretrain_recipe}a).

\smallskip\noindent\textbf{\textit{Model.}} 
The vanilla ViT~\cite{dosovitskiy2020image} is used as encoder and applied \textit{only} on the visible image patches. This design reduces time and memory complexity~\cite{he2022masked}: a masking ratio of 90\% (used in our paper) can reduce the encoder complexity to $<$1/10.
The decoder is another ViT and only used during pre-training to reconstruct the masked patches. 
Therefore, the decoder is made to be more lightweight than the encoder (depth=2, width=512). 
As a result, although the decoder processes both visible and masked image patches, its complexity is much smaller than the encoder. 
Positional embeddings are added to visible and masked patches in this full set to preserve information about their original location in the image.
We use ViT-S/16 and ViT-B/16 to denote ViT-Small and ViT-Base with a patch size of 16$\times$16 for simplicity.

\begin{table*}[!ht]
\centering
    \footnotesize

    \begin{tabular}{p{0.1\linewidth}|p{0.14\linewidth}p{0.12\linewidth}p{0.08\linewidth}|P{0.13\linewidth}P{0.13\linewidth}P{0.13\linewidth}}
        Architecture & Pre-training Dataset & Method & Annotation & ChestX-ray14 & CheXpert & COVIDx \\
        \shline
        \multirow{7}{*}{\shortstack[l]{CNN\\(DenseNet-121)}} & N/A & Random & 0 & 80.4 & 87.8 & 93.0 \\
        \cline{2-7}
         & \multirow{4}{*}{ImageNet (14M)} & Categorization & 14M & \underline{82.2} & \textbf{89.4} & 94.4 \\
         
         & & MoCo v2$^{\dagger}$~\cite{chen2020improved} & 0 & 80.9 & 87.9 & 95.5\\
         & & BYOL$^{\dagger}$~\cite{grill2020bootstrap} & 0 & 81.0 & 87.8 & 95.0 \\
         & & SwAV$^{\dagger}$~\cite{caron2020unsupervised} & 0 & 81.5 & 88.0 & \underline{95.8}\\
        \cline{2-7}
         & \multirow{2}{*}{X-rays (0.3M)} & MoCo v2~\cite{chen2020improved} & 0 & 80.6 & 88.7 & 94.0\\
         & & MAE$^{\dagger\dagger}$ & 0 & 81.2 & 88.7 & \textbf{96.5}\\
        \hline
        \multirow{4}{*}{\shortstack[l]{ViT\\(ViT-S/16)}} & N/A & Random & 0 & 67.9 & 77.9 & 87.3 \\
        \cline{2-7}
         & \multirow{2}{*}{ImageNet (14M)} & Categorization & 14M & 79.6 &88.1 & 94.3 \\
         & & MAE & 0 & 78.6 & 88.3 & 88.8 \\
        \cline{2-7}
         & X-rays (0.3M) & MAE & 0 & \textbf{82.3} & \underline{89.2} & 95.2 \\
        \hline
    \end{tabular}
    \begin{tablenotes}
        \item $^{\dagger}$The pre-trained weights of ResNet-50 were taken from Ericsson~\etal~\cite{ericsson2021well} (\textit{DenseNet is not available for advanced self-supervised ImageNet pre-training}). 
        \item $^{\dagger\dagger}$MAE was developed for ViT (not directly applicable to CNN), so we implement it based on image in-painting~\cite{pathak2016context,zhou2021models,haghighi2020learning,haghighi2021transferable}.
    \end{tablenotes}
    \caption{
        \textbf{Pre-training on ImageNet~vs.~X-rays.} 
        A direct comparison is performed between ViT and three groups of CNNs on three public datasets, considering the number of parameters, volumes of computations, usages of GPUs, and suitable pre-training schemes.
        The results suggest that ViT
        (I) consistently exceeds the CNNs that are pre-trained by state-of-the-art pre-training schemes on ImageNet, underlining the importance of in-domain transfer learning (\S\ref{sec:indomain_transfer}); 
        (II) surpasses the CNNs that are pre-trained by MAE and MoCo~v2 on the same number of medical data (0.3M X-rays); 
        (III) performs comparably (or even better) than state-of-the-art CNNs reported in the literature (detailed in Tables~\ref{tab:sota_chexpert}--\ref{tab:sota_chestxray}).
        Additionally, several important observations are obtained: 
        (\textit{i}) training from scratch takes longer epochs to converge than fine-tuning pre-trained weights (200~vs.~75 epochs);
        (\textit{ii}) ViT shows inferior performance to CNN when training from scratch on X-ray images or fine-tuning from ImageNet;
        (\textit{iii}) restorative pre-training (MAE) outperforms contrastive pre-training (MoCo~v2) in radiography imaging. 
    }
    \label{tab:comparison_init}
\end{table*}

\begin{table}[!t]
\centering
\footnotesize

\begin{tabular}{P{0.08\linewidth}|P{0.3\linewidth}P{0.25\linewidth}P{0.15\linewidth}}
                                   & RandomResizedCrop    & Crop Scale & mAUC       \\ \shline
\multirow{3}{*}{MAE}            & \xmark & N/A                & 65.9          \\
& \cmark & (0.2, 1.0)       & {69.8} \\
                                  & \cmark & (0.5, 1.0)       & \textbf{70.8} \\
\hline
\end{tabular}
\caption{
\textbf{A modest random cropping scale} is preferred for medical pre-training because the diseased regions (as foreground) is more local than photographic images, and pathological disorder could disperse over the entire X-rays~\cite{haghighi2022dira} (rather than the center of the image).
}
\label{tab:crucial_RRC}
\end{table}

\begin{table*}[t]

\centering

\subfloat[
\textbf{Layer-wise LR decay}. Learning rate decay in layer-wise needs to be tuned closely.
\label{tab:ft_lrd}
]{
\centering
\begin{minipage}{0.29\linewidth}{\begin{center}
\tablestyle{10pt}{1.2}
\begin{tabular}{cc}
Layer-wise LR decay & mAUC (\%)         \\ \shline
45                  & 82.1          \\
55                  & \textbf{82.3} \\
65                  & 82           \\
\hline
\end{tabular}
\end{center}}\end{minipage}
}
\hspace{2em}
\subfloat[
\textbf{RandAug magnitude}. A modest level of augmentation is preferred for fine-tuning.
\label{tab:ft_randaug}
]{
\begin{minipage}{0.29\linewidth}{\begin{center}
\tablestyle{10pt}{1.2}
\begin{tabular}{cc}
RandAug magnitude & mAUC (\%)          \\ \shline
4                 & 82.0          \\
6                 & \textbf{82.2} \\
8                 & 82.1          \\
\hline
\end{tabular}
\end{center}}\end{minipage}
}
\hspace{2em}
\subfloat[
\textbf{DropPath rate}. Fine-tuning on Chest X-ray images needs stronger regularization than natural images.
\label{tab:mask_token}
]{
\begin{minipage}{0.29\linewidth}{\begin{center}
\tablestyle{10pt}{1.2}
\begin{tabular}{cc}
DropPath rate & mAUC (\%)          \\ \shline
0.1           & 81.5          \\
0.2           & \textbf{82.3} \\
0.3           & 82.1           \\
\hline
\end{tabular}
\end{center}}\end{minipage}
}
\\
\centering
\caption{\textbf{The fine-tuning recipe.} The ablation studies are conducted using ViT-S/16 on NIH Chest X-ray14. We report the 14-class average AUC (\%). Except for (b) using Layer-wise LR decay 0.65, all the experiments adopt the optimal value for the hyper-parameters (Layer-wise LR decay 0.55, RandAug magnitude 6, and DropPath rate 0.2). }
\label{tab:ablations_ft_recipe} 
\end{table*}

\section{Pre-training: Recipe and Results}
\label{sec:pretraining_recipe_result}

\noindent\textbf{\textit{Implementation details.}}
We adopt AdamW~\cite{loshchilov2018decoupled} optimizer with $\beta_1=0.9, \beta_1=0.95$ and set the weight decay to 0.05. 
Transformer blocks are initialized with xavier\_uniform~\cite{glorot2010understanding}. 
We set learning rate ($lr$) and batch size to 1.5$e$-4 and 2048. 
$lr$ is warmed up for 20 epochs~\cite{goyal2017accurate} and scheduled with cosine annealing strategy~\cite{loshchilov2016sgdr}.
The pre-training stage takes 800 epochs in total.
Random resized crop and horizontal flip are used as data augmentation.

\subsection{On the importance of in-domain transfer}
\label{sec:indomain_transfer}

\tableautorefname~\ref{tab:comparison_init} provides a comprehensive comparison on three sets of model initialization: random, ImageNet pre-training, and X-ray pre-training.
Unlike CNNs, ViTs trained from scratch show very poor performance even with a strong training recipe and 2.7$\times$ larger number of training epochs than fine-tuning. 
On the contrary, ViTs achieve comparable and sometimes superior performance with the help of pre-training on large-scale datasets (\eg ImageNet and X-rays). 
Specifically, after supervised pre-trained on ImageNet (following~\cite{touvron2021training}), ViT-S/16 shows acceptable performance on the three datasets but is still distanced to the state-of-the-art CNN-based methods.
In-domain transfer seeks to reduce domain disparities between photographic and medical images~\cite{hosseinzadeh2021systematic}.
In doing so, we bridge the domain gap and satisfy the ViTs/CNNs appetite for data by pre-training on 0.3M unlabeled chest X-rays.
ViTs benefit more on the in-domain transfer (improved mAUC from 78.6 to 82.3 on ChestX-ray14), whereas ImageNet pre-trained CNNs remain high performance compared with in-domain pre-training (82.2~vs.~81.2 on ChestX-ray14).

\subsection{Learning from 266,340 unlabeled X-rays}
\label{sec:more_data}

Training ViTs from scratch is harder than CNNs because ViTs lack inductive bias in modeling local visual representation and generally require more data to figure out the image content on their own~\cite{steiner2021train,liu2022convnet,matsoukas2021time}. As shown in~\tableautorefname~\ref{tab:comparison_init}, supervised pre-training on ImageNet brings performance gain from 67.9\% to 79.6\% for ViTs, and from 80.4\% to 82.1\% for CNNs on ChestX-ray14.
We ask: \textit{How many X-rays are needed for ViT pre-training?} 
\figureautorefname~\ref{fig:pretrain_recipe}a shows that ViT-S/16 pre-trained on 75K, 191K, 266K, and 510K X-rays achieve a mAUC of 79.3\%, 81.9\%, 82.3\%, and 82.3\% on ChestX-ray14. 
The improvement from 75K to 266K is statistically significant ($p$-value=1.2$e$-127), but the performance gain is negligible from 266K to 510K---a bottleneck for ViT-S/16 (with 22M parameters). Although larger ViTs (\eg ViT-B/16 with 86M parameters) can produce higher performance, considering computational cost and the fairness of the ViT~vs.~CNN, we end up pre-training ViT-S/16 using 266,340 unlabeled X-rays for benchmarking.

\subsection{Masking out 90\% X-ray content}
\label{sec:more_mask}

The optimal masking ratio is related to the information redundancy in the data: BERT~\cite{kenton2019bert} uses a masking ratio of 15\% for language and MAE~\cite{he2022masked} uses a ratio of 75\% for images. 
Most recent study suggests that videos, due to its greater redundancy in the temporal dimension, can apply only 90\% masking ratio for pre-training~\cite{feichtenhofer2022masked}.
Given the great similarity in the chest anatomy, naturally, we hypothesize that \textit{chest X-rays require even larger masking ratio for pre-training.}
This is in line with the assumption that chest X-rays are more information-redundant than photographic images.
We experimented with masking ratios ranging from 75\% to 95\%, incremented by intervals of 5\%. \figureautorefname~\ref{fig:pretrain_recipe}b indicates that 90\% is the optimal masking ratio for MAE pre-training on chest X-rays. The larger masking ratio results in a more efficient pre-training, which is 2.5$\times$ faster than the original MAE. The efficient pre-training, in turn, enables us to scale up to larger ViT architectures and more diverse datasets.

\subsection{Cropping patches scaled of (0.5$\sim$1.0)}
\label{sec:less_cropping}

As the spatial consistency of medical images is much higher than photographic images, there is a need to analyze the effectiveness of spatial data augmentation (\eg random resized crop). 
\figureautorefname~\ref{fig:pretrain_recipe}b suggests that RandomResizedCrop operation has consistent and noticeable benefit to MAE pre-training on chest X-ray imaging under different masking ratios.
It enables ViTs to learn multiscale features from X-rays and to avoid the over-fitting problem due to the lack of training examples.
More importantly, a relatively smaller cropping ratio than natural imaging is preferred (\tableautorefname~\ref{tab:crucial_RRC}). Cropping patches scaled of (0.5$\sim$1.0) yields 1.0\% higher mAUC than those of (0.2$\sim$1.0) (as suggested in~\cite{he2022masked}). It is intuitive that strong spatial augmentation is harmful since the informative lesions or organs could be cropped and biased and models will be learned with noisy annotations.

\begin{figure*}[!t]
    \centering
    \includegraphics[width=1.0\linewidth]{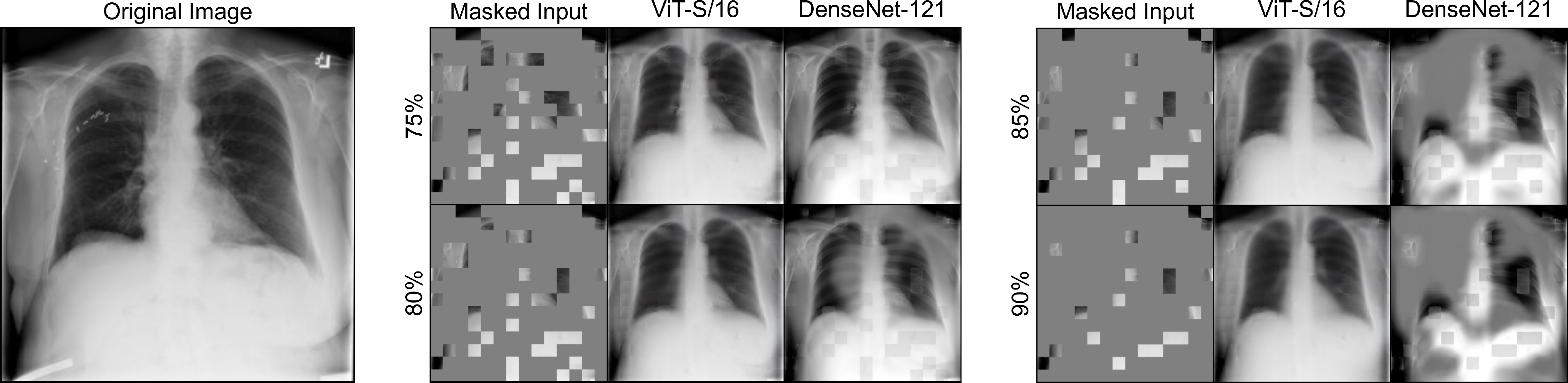}
    \caption{
    \textbf{Reconstruction of ChestX-ray14 \textit{validation} images.} Pre-trained with a masking ratio of 75\%, ViTs generalizes better than CNNs to the input images that are applied with higher masking ratios.
    }
    \label{fig:reconstruction}
    % \vspace{-3mm}
\end{figure*}

\subsection{Quality assessment of image reconstruction}
\label{sec:quality_assessment_reconstruction}

We assess reconstruction quality using validation images in \figureautorefname~\ref{fig:reconstruction} for both ViT and CNN. 
The models are pre-trained on ChestX-ray14 and evaluated on the inputs with varying masking ratios, spanning from 75\% to 90\%, incremented by intervals of 5\%. 
Both ViTs and CNNs can predict the overall anatomical structures in X-rays, but fail to reconstruct detailed texture such as shoulder bones.
This is expected because the ViTs/CNNs only see 10\% of the input image and attempt to reconstruct the rest 90\% during the training---it is difficult even for expert radiologists.
There is no clear evidence showing that the reconstruction capability is positively correlated to the transfer learning performance.
On the contrary, the original autoencoders~\cite{hinton2006reducing} (with a masking ratio of 0\%) can certainly reconstruct images better than \textit{masked} autoencoders, but their resulting representation is not as effective as the masked counterparts~\cite{zhou2021models}. 
Moreover, studies in both CNNs~\cite{tao2020revisiting,zhou2021models} and ViTs~\cite{he2022masked,xie2022simmim} indicate that alternative loss functions (\eg $l$1,  smooth-$l$1, SSIM, and adversarial losses) for reconstruction would not contribute to the transfer learning performance. Therefore, we used $l$2 loss as default.
Finally, we should remark that our ultimate goal is not the task of image reconstruction \textit{per se}. While reconstructing patches is advocated and investigated as a pre-training scheme for ViTs/CNNs, the usefulness of the learned representation must be assessed objectively based on its generalizability and transferability to various downstream tasks (presented in~\S\ref{sec:finetuning_recipe_results}).
\section{Fine-tuning: Recipe and Results}
\label{sec:finetuning_recipe_results}

\begin{table*}[!ht]
\centering
    \footnotesize
    \begin{tabular}{p{0.05\linewidth}p{0.12\linewidth}P{0.14\linewidth}P{0.07\linewidth}P{0.09\linewidth}P{0.09\linewidth}P{0.05\linewidth}P{0.06\linewidth}P{0.09\linewidth}}
        \multicolumn{2}{l}{Method}  & Architecture & Atelectasis & Cardiomegaly & Consolidation & Edema & Effusion & mAUC (\%)\\
        \shline
        \multicolumn{2}{l}{Allaouzi~\etal~\cite{allaouzi2019novel}} & \multirow{7}{*}{DN121} & 72.0 & \textbf{88.0} & 77.0 & 87.0 & 90.0 & 82.8 \\
        \multicolumn{2}{l}{Irvin~\etal~\cite{irvin2019chexpert}} &  & 81.8 & 82.8 & \underline{93.8} & \underline{93.4} & 92.8 & 88.9 \\
        \multicolumn{2}{l}{Seyyedkalantari~\etal~\cite{seyyedkalantari2020chexclusion}} &  & 81.2 & 83.0 & 90.0 & 88.3 & \textbf{93.8} & 87.3 \\
        \multicolumn{2}{l}{Pham~\etal~\cite{pham2021interpreting}}  &  & \underline{82.5} & 85.5 & 93.7 & 93.0 & 92.3 & \textbf{89.4} \\
        \multicolumn{2}{l}{Hosseinzadeh~\etal~\cite{hosseinzadeh2021systematic}} &  & - & - & - & - & - & 87.1 \\
        \multicolumn{2}{l}{Haghighi~\etal~\cite{haghighi2022dira}} &  & - & - & - & - & - & 87.6 \\
        \multicolumn{2}{l}{Kang~\etal~\cite{kang2021data}} &  & 82.1 & \underline{85.9} & \textbf{94.4} & 89.2 & \underline{93.6} & 89.0 \\
        \hline
        \multirow{4}{*}{Ours} & MoCo v2 & DN121 &  78.5 & 77.9 & 92.5 & 92.8 & 92.7 & 88.7\\
         & MAE & DN121 & 81.5 & 77.6 & 89.4 & 92.3 & 92.0 & 88.7\\
         & MAE & ViT-S/16 & \textbf{83.5} & 81.8 & 93.5 & \textbf{94.0} & 93.2 & \underline{89.2} \\
        \cline{2-9}
         & \gc{MAE} & \gc{ViT-B/16} & \gc{82.7} & \gc{83.5} & \gc{92.5} & \gc{93.8} & \gc{94.1} & \gc{89.3} \\
        \hline
    \end{tabular}

    \caption{
        \textbf{CheXpert benchmark.} ViT achieves comparable performance to the state-of-the-art CNNs on CheXpert (\textit{official val}) over all five thorax diseases and the best o ``Atelectasis'' and ``Edema'' diseases.
    }
    \label{tab:sota_chexpert}
\end{table*}

\begin{table*}[!ht]
    \centering
    \footnotesize
    \begin{tabular}{p{0.05\linewidth}p{0.08\linewidth}p{0.08\linewidth}P{0.15\linewidth}P{0.1\linewidth}P{0.08\linewidth}P{0.09\linewidth}P{0.15\linewidth}}

        \multicolumn{3}{l}{Method} & Input Resolution & \# Params (M) & MACs (G) & Accuracy & COVID-19 Sensitivity \\
        \shline
        \multicolumn{3}{l}{COVIDNet-CXR-3} & \multirow{8}{*}{480$\times$480} & 29  & 29.1  & 98.3 & 97.5 \\ 
        \multicolumn{3}{l}{COVIDNet-CXR-2} &  & 9  & 5.6  & 96.3 & 95.5 \\ 
        \multicolumn{3}{l}{COVIDNet-CXR4-A} &  & 40  & 23.6  & 94.3 & 95.0 \\ 
        \multicolumn{3}{l}{COVIDNet-CXR4-B} &  & 12  & 7.5  & 93.7 & 93.0 \\ 
        \multicolumn{3}{l}{COVIDNet-CXR4-C} &  & 9  & 5.6  & 93.3 & 96.0 \\ 
        \multicolumn{3}{l}{COVIDNet-CXR3-A} &  & 40  & 23.6  & 93.3 & 94.0 \\ 
        \multicolumn{3}{l}{COVIDNet-CXR3-B} &  & 12  & 7.5  & 93.3 & 91.0 \\ 
        \multicolumn{3}{l}{COVIDNet-CXR3-C} &  & 9  & 5.6  & 92.3 & 95.0 \\
        \hline
        \multirow{4}{*}{Ours} & MoCo~v2 & DN121 & \multirow{3}{*}{448$\times$448} &7 & 11.6 & 96.0 & 96.5\\
         & MAE & DN121 &  &7 & 11.6 & 96.3 & 98.0\\
         & MAE & ViT-S/16 &  & 22 & 16.9 & 95.3 & 95.0 \\
        \cline{2-8}
         & \gc{MAE} & \gc{ViT-B/16} & \gc{448$\times$448} & \gc{86} & \gc{67.2} & \gc{97.3} & \gc{98.0} \\
        \hline\hline
        \multicolumn{3}{l}{COVIDNet-CXR Small} & \multirow{2}{*}{224$\times$224} & 117  & 2.3  & 92.6 & $\:\:$87.1$^{\dagger}$\\ 
        \multicolumn{3}{l}{COVIDNet-CXR Large} &  & 127  & 3.6  & 94.4 & $\:\:$96.8$^{\dagger}$\\ 
        \hline
        \multirow{4}{*}{Ours} & MoCo~v2 & DN121 & \multirow{3}{*}{224$\times$224} &7 & 2.9 & 94.0 & 94.5 \\
         & MAE & DN121 &  &7 & 2.9 & \textbf{96.5} & 97.0\\
         & MAE & ViT-S/16 &  & 22 & 4.2 & \underline{95.2} & 94.5 \\
        \cline{2-8}
         & \gc{MAE} & \gc{ViT-B/16} & \gc{224$\times$224} & \gc{86} & \gc{16.9} & \gc{95.3} & \gc{95.5} \\ 
        \hline
    \end{tabular}
    \begin{tablenotes}
        \item $^{\dagger}$The results are evaluated on 31 images; otherwise, the results are evaluated on the latest official testing set (400 images).
    \end{tablenotes}
    \caption{
        {\jlmagenta
        \textbf{COVIDx benchmark.} ViTs show comparable performance to state-of-the-art CNNs on COVIDx (\textit{official val}).}
    }
    \label{tab:sota_covidx}
\end{table*}

\begin{table}[!ht]
\centering
    \footnotesize

    \begin{tabular}{p{0.1\linewidth}p{0.14\linewidth}P{0.19\linewidth}P{0.23\linewidth}P{0.08\linewidth}}
        \multicolumn{2}{l}{Method} & Architecture & Pre-training  & mAUC \\
        \shline
        \multicolumn{2}{l}{Wang~\etal~\cite{wang2017chestx}} & RN50 & \multirow{16}{*}{ImageNet (14M)} & 74.5 \\
        \multicolumn{2}{l}{Yao~\etal~\cite{yao2018weakly}} & RN\&DN &  & 76.1\\
        \multicolumn{2}{l}{Li~\etal~\cite{li2018thoracic}} & RN50 &  & 75.5 \\
        \multicolumn{2}{l}{Tang~\etal~\cite{tang2018attention}} & RN50 &  & 80.3\\
        \multicolumn{2}{l}{Guendel~\etal~\cite{guendel2018learning}} & DN121 &  & 80.7\\
        \multicolumn{2}{l}{Guan~\etal~\cite{guan2018multi}} & DN121 &  &81.6\\
        \multicolumn{2}{l}{Wang~\etal~\cite{wang2019thorax}} & R152 &   & 78.8 \\
        \multicolumn{2}{l}{Ma~\etal~\cite{ma2019multi}} & R101  &  & 79.4\\
        \multicolumn{2}{l}{Baltruschat~\etal~\cite{baltruschat2019comparison}} & RN50 &  & 80.6\\
        \multicolumn{2}{l}{Seyyed~\etal~\cite{seyyedkalantari2020chexclusion}} &  DN121 &  &81.2 \\
        \multicolumn{2}{l}{Ma~\etal~\cite{ma2020multilabel}} & DN121($\times$2) &  & 81.7\\
        \multicolumn{2}{l}{Hermoza~\etal~\cite{hermoza2020region}} & DN121 &  & 82.1\\
        \multicolumn{2}{l}{Kim~\etal~\cite{Kim_2021_CVPR}} & DN121 &  & \underline{82.2}\\
        \multicolumn{2}{l}{Haghighi~\etal~\cite{haghighi2022dira}} & DN121 &  & 81.7 \\
        \multicolumn{2}{l}{Liu~\etal~\cite{liu2022acpl}} & DN121 &  & 81.8 \\
        \multicolumn{2}{l}{Taslimi~\etal\cite{taslimi2022swinchex}} & SwinT &  & 81.0 \\
        \hline
        \multirow{4}{*}{Ours} & MoCo v2 & DN121 & \multirow{3}{*}{X-rays (0.3M)} &  80.6\\
         & MAE & DN121 &  & 81.2 \\
         & MAE & ViT-S/16 &  & \textbf{82.3} \\
         \cline{2-5}
         & \gc{MAE} & \gc{ViT-B/16} & \gc{X-rays (0.5M)} & \gc{83.0} \\
        \hline
    \end{tabular}
    \caption{
        \textbf{ChestX-ray14 benchmark.}
        ViT-S/16 achieves comparable performance to previous state-of-the-art CNN-based and Transformer-based methods on ChestX-ray14 (\emph{official} split) reported in the literature. With the same pre-training scheme (MAE) on 0.3M X-rays, ViT significantly outperforms its CNN counterparts. In addition, ViT-B/16, pre-trained on 0.5M X-rays, hits a new record of 83.0 mAUC. RN, DN, and SwinT denote ResNet, DenseNet, and Swin Transformer.
    }
    \label{tab:sota_chestxray}
\end{table}

\noindent\textbf{\textit{Fine-tuning.}} The optimizer and $lr$ scheduler are the same as pre-training. The choices of layer-wise LR decay, RandAug~\cite{Cubuk2019} magnitude\footnote{No improvement is obtained by more aggressive augmentation strategies (\ie mixup~\cite{zhang2018mixup} and cutmix~\cite{yun2019cutmix}) since they could produce noisy labels by removing or overlapping thorax diseases in X-rays.}, and DropPath~\cite{huang2016deep} rate are crucial to fine-tune the pre-trained ViTs. The optimal settings are given by extensive studies in~\tableautorefname~\ref{tab:ablations_ft_recipe} and we reuse them for all three radiography imaging tasks in~\S\ref{sec:chestxray14_result}--\ref{sec:covidx_result}. Models are fine-tuned with 75 epochs on all three datasets. 

\smallskip\noindent\textbf{\textit{Linear-probing.}} LARS~\cite{you2017large} optimizer is used with \textit{momentum}=0.9. 
We set learning rate ($lr$) and batch size to 0.1 and 16,384. 
$lr$ is warmed up~\cite{goyal2017accurate} for 10 epochs and scheduled with cosine annealing strategy~\cite{loshchilov2016sgdr}.
The ViT is trained with 100 epochs. Linear-probing is used in~\figureautorefname~\ref{fig:pretrain_recipe}b.

\subsection{Stanford CheXpert}
\label{sec:chexpert_result}

\noindent\textbf{\textit{Experimental setup.}}
CheXpert is a large scale dataset containing 191,028 frontal-view chest X-rays. 14 diseases in radiology reports exist in the dataset and five common diseases are for benchmarking. We resized the images into 224$\times$224 and the test is done on the official validation set.
Mean Area Under the Curve (AUC) on five classes is reported for comparison.

\smallskip\noindent\textbf{\textit{Results and analysis.}}
As shown in \Cref{tab:sota_chexpert}, vanilla ViT-S achieves 89.2\% mAUC which is very competitive to the best performance of 89.4\%. Moreover, ViT-S yields the best performance on diseases of Atel (83.5\%) and Edem (94.0\%).

\subsection{COVIDx}
\label{sec:covidx_result}

\noindent\textbf{\textit{Experimental setup.}}
COVIDx (version 9A) provides over 30,000 images containing 16,490 positive COVID-19 images. The dataset is annotated with 4 different classes for the training set of 30,130 images while the testing set only has 400 images of 3 classes. To ensure a fair comparison with previous methods, Accuracy and COVID-19 sensitivity on the testing set (3 classes) are reported. 

\smallskip\noindent\textbf{\textit{Results and analysis.}} We compare our vanilla ViT-S model with the state-of-the-art models provided on the official github repository\footnote{\href{https://github.com/lindawangg/COVID-Net/blob/master/docs/models.md}{github.com/lindawangg/COVID-Net/blob/master/docs/models.md}}. Our method beats the two other models when input resolution is 224$\times$224 while achieving a very high accuracy of 95.2\% and COVID-19 sensitivity of 94.5\%. ViT-S shows a great balance between the model size, computation cost, and the performance.

\subsection{NIH ChestX-ray14}
\label{sec:chestxray14_result}

\noindent\textbf{\textit{Experimental setup.}}
ChestX-ray14 has 112,120 frontal-view X-rays of 30,805 unique patients with the text-mined fourteen disease labels (where each image can have multiple labels). We follow the official data split which assigns 75,312 images for training and 25,596 images for testing. We resize the original images from size 1024$\times$1024 into 224$\times$224. Mean AUC on 14 classes is reported and 17 most popular and compelling baseline methods are compared.

\smallskip\noindent\textbf{\textit{Results and analysis.}} \Cref{tab:sota_chestxray} provides a systematic comparison with state-of-the-art CNN and Transformer models on NIH ChestX-ray14 over years. The previous best CNN performance was obtained by DenseNet-121~\cite{wang2021triple} with a mean AUC of 82.6\%. The previous best performance of Transformers was 81.0\%~\cite{taslimi2022swinchex}, which was distanced to CNN's performance. Our vanilla ViT-S shows a very competitive result of 82.3\% mean AUC over 14 diseases with the best classification performance on 6 out of 14 thorax diseases. It is worth noting that the research community took four years to improve the AUC score from 74.5 to 82.2 for CNN-type architectures, largely due to the difficulty of the training recipe.

\section{Discussion}
 
\label{sec:discussion}

\begin{figure}[!t]
    \centering
    \includegraphics[width=1.0\linewidth]{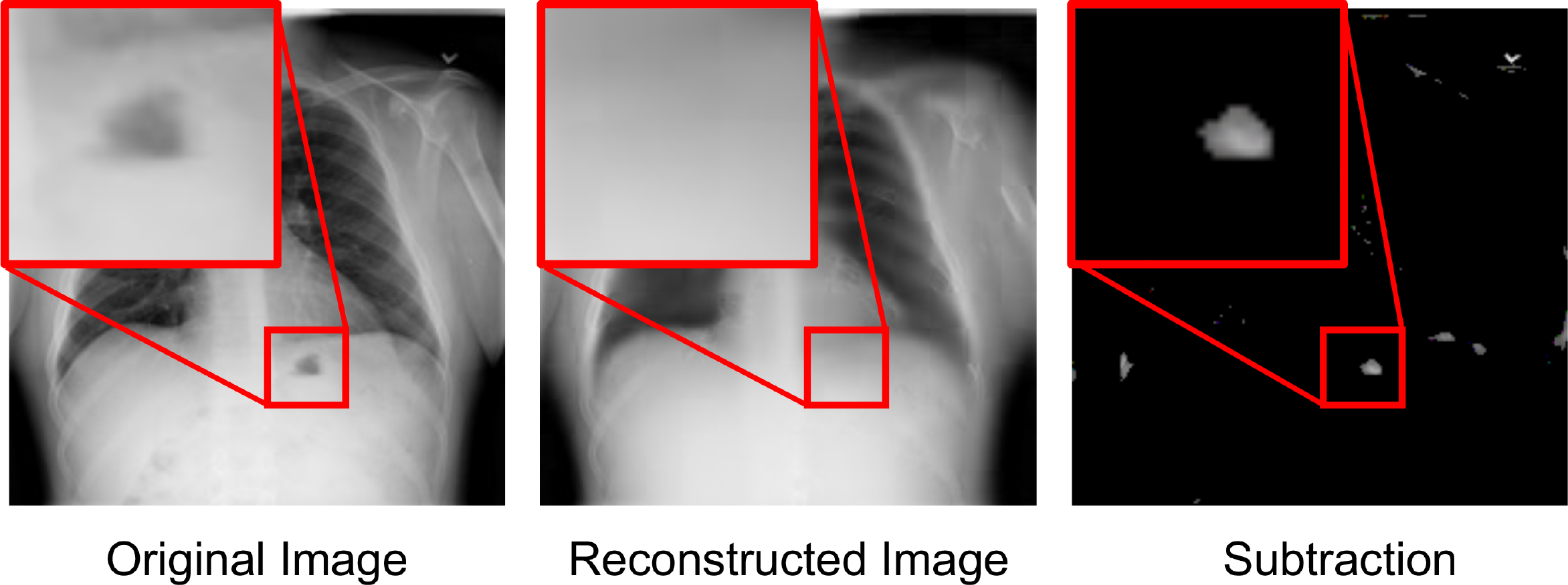}
    \caption{\textbf{MAE could reveal anomalies.}
    We input a chest X-ray (with anomalies) to the trained MAE and plot the difference map of the reconstructed output and original X-ray. 
    Interestingly, we observe that the MAE happens to ``heal'' those anomalies by replacing them with normal patterns. }
    
    \label{fig:anomaly_detection}
\end{figure}

\noindent\textbf{\textit{Can MAE detect anomalies from an image?}}
Anomalies are something that appear differently from the normal X-rays---can be diseases, medical devices, and clinical notations (\eg arrows, numbers, letters).
Since MAE is trained using the original X-rays as ground truth and the majority pixels in an X-ray are normal, the MAE should be able to overfit the normal anatomical patterns. 
Now, \textit{if an anomaly is masked out, can MAE reconstruct a normal pattern?}
If so, by subtracting the reconstructed output and the original X-ray, the anomaly can be detected and localized.
A similar point has been discussed in Zhou~\etal~\cite{zhou2021models}, but was illustrated using CNNs.
Specifically, we input an original X-ray to the trained MAE and plot the difference map between reconstructed output and the original image.
As shown in~\figureautorefname~\ref{fig:anomaly_detection}, MAE happens to ``heal'' those anomalies and reconstruct with normal patterns. 
This behavior can be thought of as an attempt to detect and localize anomalies. 
More importantly, unlike weakly-supervised detection strategies~\cite{zhou2016learning,baumgartner2018visual,cai2018iterative,siddiquee2019learning,xiang2021painting}, neither image-level nor pixel-level annotation is required for this approach, making it an attractive and challenging direction to explore~\cite{pinaya2022unsupervised,tian2022unsupervised}.

\begin{table}[!t]
\centering
    \footnotesize
    \begin{tabular}{p{0.2\linewidth}P{0.13\linewidth}|P{0.09\linewidth}P{0.09\linewidth}|P{0.09\linewidth}P{0.09\linewidth}}
        % \hline
         & Size & \multicolumn{2}{c}{DenseNet-121} & \multicolumn{2}{c}{ViT-S/16} \\
        Disease & (\# of px) & AP$_{25}$ & AP$_{50}$ & AP$_{25}$ & AP$_{50}$ \\
        \shline
        Nodule & 224 & 0.0 & 0.0 & 9.2 & 3.9 \\
        Mass & 756 & 25.4 & 1.6 & 27.0 & 11.1 \\
        Atelectasis & 924 & 10.1 & 2.0 & 31.5 & 8.1 \\
        Pneumothorax & 1899 & 11.6 & 2.3 & 4.7 & 0.0 \\
        Infiltrate & 2754 & 32.9 & 12.7 & 11.4 & 1.3 \\
        Effusion & 2925 & 24.5 & 2.9 & 8.8 & 1.0 \\
        Pneumonia & 2944 & 32.0 & 6.2 & 27.8 & 9.3 \\
        Cardiomegaly & 8670 & 89.6 & 53.3 & 16.3 & 3.0 \\ 
        \hline
        All & 2300 & 31.0 & 12.3 & 18.0 & 4.7 \\
        \hline
    \end{tabular}

    \caption{
        \textbf{Weakly-supervised disease localization.} 
        We report the average precision (AP) on 25\% and 50\% IoUs. The IoU is calculated between the ground truth bounding box and bounding box of the largest connected component in the Grad-CAM heatmap.
        We also present the statistics of disease sizes, measured by the number of pixels within the bounding box, showing that CNN can detect large diseases (\eg Cardiomegaly, Pneumonia) better than ViT, while ViT can capture smaller diseases (\eg nodule).
    }
    \label{tab:appendix_detection_results}
\end{table}

\begin{figure}[!t]
    \centering
    \includegraphics[width=1.0\linewidth]{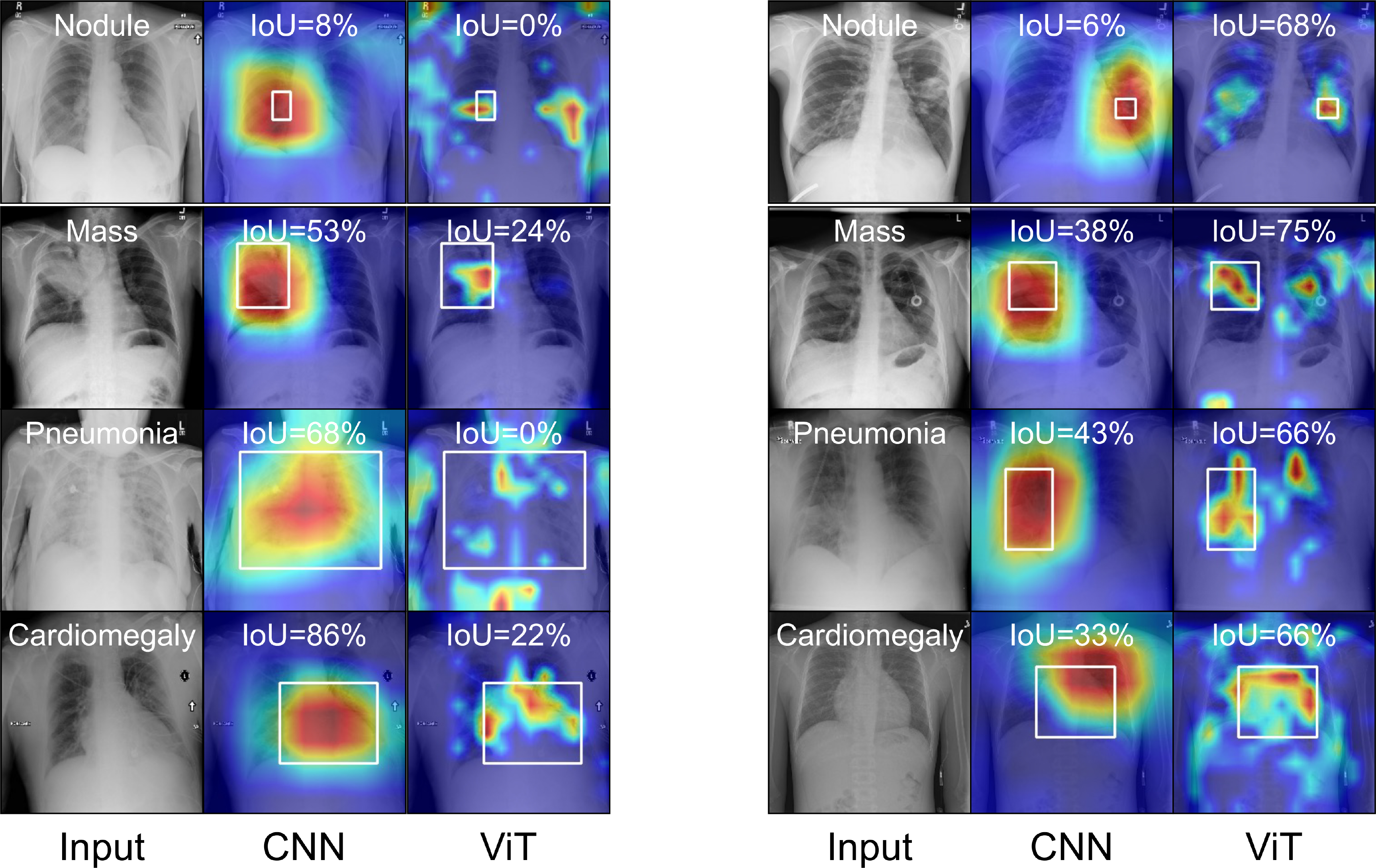}
    \caption{\textbf{Grad-CAM of CNN and ViT.} [Better viewed on-line, in color, and zoomed in for details] ChestX-ray14 provides bounding boxes for some of the thorax diseases, shown in white boxes.
    Left and right panels display successful cases predicted by CNN and ViT, respectively. 
    }
    \label{fig:grad_cam_small}
\end{figure}

\smallskip\noindent\textbf{\textit{Weakly-supervised disease localization by ViT and CNN.}}
With the help of Grad-CAM\footnote{\href{https://github.com/jacobgil/pytorch-grad-cam}{github.com/jacobgil/pytorch-grad-cam}}~\cite{jacobgilpytorchcam}, we are able to check which part of the X-ray image is responsible for the model prediction (the diseased region). We use the last dense-block (4th) of DenseNet-121 and the LayerNorm layer in the last transformer block (12th) of ViT-S/16 as the ``target layers'' for Grad-CAM. 
The experiments are done in a small subset of ChestX-ray14, which offers 787 cases with bounding-box of a total of eight thorax diseases. The final predicted bounding-box of the diseased region is generated with the thresholded Grad-CAM heatmap, largest connected component, and box regression.
The results are evaluated by IoU between ground truth bounding box and the bounding box of the largest connected component in the attention response. 
We then compute Average Precision (AP) as the detection metric~\cite{lin2014microsoft}. 
Precision is defined as $tp/(tp+fp)$, where $tp$ and $fp$ denote the number of true positives and false positives, respectively.
AP$_{25}$ considers cases with IoU$>$25\% as true positives and $AP_{50}$ with IoU$>$50\%. \tableautorefname~\ref{tab:appendix_detection_results} shows the detection results (including disease-wise results and all diseases). 
We observe that the CNN provides better localization of diseases in a larger size (\eg Cardiomegaly and Pneumonia) while ViT is robust to diseases in a smaller size (\eg Nodule). 
Although the classification performance of CNN and ViT is comparable (82.1\%~vs.~82.3\% AUC), CNN significantly exceeds the localization ability to ViT, and their attention maps generated by Grad-CAM behave differently.
\figureautorefname~\ref{fig:grad_cam_small} provides examples of GradCAM of CNN and ViT.
Attentions in CNN are relatively larger and more concentrated than those in ViT.
This observation is consistent with those in Chefer~\etal~\cite{chefer2021transformer}.
This study suggests that class activation maps are more suitable for visualizing the explainability of CNN-type models. In the future, other than class activation maps, we will seek to explore the explainability for Vision Transformers in multi-label classification tasks, with the help of self-attention derived from the Transformer architectures~\cite{caron2021emerging,raghu2021vision,abnar2020quantifying,chefer2021transformer}.

\section{Conclusion and Future Work}
\label{sec:conclusion_future}

This paper has unleashed the potential of stand-alone, vanilla ViT by devising strong pre-training and fine-tuning recipes. 
We overcome several technical barriers and bring reusable insights for the medical vision community.
Specifically, we (\textit{i}) improve computational efficiency; (\textit{ii}) customize data augmentation; (\textit{iii}) explore larger data scale; and (\textit{iv}) optimize learning parameters.
As a result, the vanilla ViT achieves a comparable (sometimes better) performance to state-of-the-art CNNs. Code and pre-trained models are available.

This paper has also presented an up-to-date benchmark on three predominant chest X-ray datasets.
Taking into account the number of parameters, volumes of computations, usages of GPUs, and suitable pre-training schemes, we have performed a fair and comprehensive comparison between vanilla ViT and (\textit{i}) state-of-the-art CNNs reported in the literature, (\textit{ii}) CNNs that are pre-trained by advanced pre-training schemes on ImageNet, (\textit{iii}) CNNs that are pre-trained on the same number of medical data. We hope this study can direct future research on the application of Transformers to a larger variety of medical imaging tasks.

As future work, we will consider three extensions to our current study. 
First, assembling more publicly available X-ray datasets for pre-training (which account for a total of $\sim$1M images~\cite{ccalli2021deep}). Scaling up the data is perhaps the most straightforward way to enhance larger ViTs (\eg ViT-Large, ViT/Huge) in terms of performance and generalizability based on \figureautorefname~\ref{fig:pretrain_recipe}a. 
Second, extending ViT to its 3D form for higher dimensional medical modalities (\eg CT, MRI), which is expected to take a considerable computational resource~\cite{hatamizadeh2022unetr} and larger data for pre-training~\cite{tang2022self}, therefore requiring a more efficient method.
Third, exploiting paired information of radiology reports and image data for pre-training. We acknowledge the unique ability of Transformers in processing multi-modality data~\cite{radford2021learning,chen2022multi}.

\smallskip\noindent\textbf{Acknowledgements.} This work was supported by the Lustgarten Foundation for Pancreatic Cancer Research.
We thank Y.~Zhang for providing data loader of the COVIDx dataset; A.~Delaney for improving the writing of this paper.

\newpage
{\small
\bibliographystyle{ieee_fullname}
\bibliography{refs}
}

\end{document}